\pdfoutput=1
\documentclass[11pt]{article}

\usepackage[final]{acl}

\usepackage{times}
\usepackage{latexsym}
\usepackage{booktabs}
\usepackage{amsfonts}  % OR \usepackage{amssymb}
 \usepackage{amsmath}

\usepackage[T1]{fontenc}
\usepackage[utf8]{inputenc}

\usepackage{microtype}

\usepackage{inconsolata}

\usepackage{graphicx}

\title{RePanda: Pandas-powered Tabular Verification and Reasoning}

\author{
 \textbf{Atoosa Malemir Chegini\textsuperscript{1}},
 \textbf{Keivan Rezaei\textsuperscript{1}},
 \textbf{Hamid Eghbalzadeh\textsuperscript{2}},
 \textbf{Soheil Feizi\textsuperscript{1}},
\\
\\
 \textsuperscript{1}Department of Computer Science, University of Maryland,
 \textsuperscript{2}AI at Meta
\\
 \small{
   \textbf{Correspondence:} \href{mailto:atoocheg@umd.edu}{atoocheg@umd.edu}
 }
}
\newcommand{\methodname}{RePanda}
\newcommand{\tabfact}{PanTabFact}
\newcommand{\wikitable}{PanWiki}
\newcommand{\wikifact}{WikiFact}

\begin{document}
\maketitle
\begin{abstract}
Fact-checking tabular data is essential for ensuring the accuracy of structured information. However, existing methods often rely on black-box models with opaque reasoning. We introduce \textit{\methodname}, a structured fact verification approach that translates claims into executable \texttt{pandas} queries, enabling interpretable and verifiable reasoning.

To train \textit{\methodname}, we construct \textit{\tabfact}, a structured dataset derived from the TabFact train set, where claims are paired with executable queries generated using DeepSeek-Chat \footnote{All experiments, data collection, and processing activities were conducted by the University of Maryland, College Park (UMD). Meta was involved solely in an advisory role and no experiments, data collection or processing activities were conducted using Meta tools or within its IT environment.} and refined through automated error correction. Fine-tuning DeepSeek-coder-7B-instruct-v1.5 on \textit{\tabfact}, \textit{\methodname} achieves 84.09\% accuracy on the TabFact test set. To evaluate Out-of-Distribution (OOD) generalization, we interpret question-answer pairs from WikiTableQuestions as factual claims and refer to this dataset as \textit{\wikifact}. Without additional fine-tuning, \textit{\methodname} achieves 84.72\% accuracy on \textit{\wikifact}, significantly outperforming all other baselines and demonstrating strong OOD robustness. Notably, these results closely match the zero-shot performance of DeepSeek-Chat (671B), indicating that our fine-tuning approach effectively distills structured reasoning from a much larger model into a compact, locally executable 7B model.

Beyond fact verification, \textit{\methodname} extends to tabular question answering by generating executable queries that retrieve precise answers. To support this, we introduce \textit{\wikitable}, a dataset mapping WikiTableQuestions to \texttt{pandas} queries. Fine-tuning on \textit{\wikitable}, \textit{\methodname} achieves 75.1\% accuracy in direct answer retrieval. These results highlight the effectiveness of structured execution-based reasoning for tabular verification and question answering. We have publicly released the dataset on Hugging Face at \href{https://huggingface.co/datasets/AtoosaChegini/PanTabFact}{datasets/AtoosaChegini/PanTabFact}.
\end{abstract}

\section{Introduction}

Fact verification is a critical task in artificial intelligence, with applications in journalism, financial auditing, and scientific research. As misinformation continues to proliferate, automated fact-checking has become an essential tool for verifying claims against structured sources such as tabular data. Unlike textual fact verification, which matches claims against unstructured text, tabular fact verification requires reasoning over structured numerical and categorical data, making it a fundamentally different and more challenging task\cite{herzig2020tapas, gu2022pasta}.

Despite recent progress in large language models (LLMs), their ability to reason over structured tabular data remains limited. LLMs are pre-trained predominantly on unstructured text, where semantic relationships are inferred through sequential token dependencies. However, tabular data encodes information in a structured format where relationships are often implicit, requiring operations such as aggregation, filtering, and comparison across multiple rows and columns~\cite{liu2021tapex, eisenschlos2020understanding}. 

A major challenge lies in LLMs' limited understanding of structured data, such as tables, when processing tabular information. Models like TAPAS~\cite{herzig2020tapas} and TAPEX~\cite{liu2021tapex} attempt to address this by incorporating table-aware pretraining to improve table understanding. However, these approaches flatten tables into sequences, losing structural integrity and making it difficult to capture relationships between rows and columns~\cite{gu2022pasta}. As a result, they struggle with complex table operations, such as aggregation and multi-row comparisons, which are essential for fact verification. Moreover, approaches like PASTA~\cite{gu2022pasta} use sentence-table cloze pretraining to improve table understanding but rely on predefined operations, which may not generalize to complex, unseen queries.

Another fundamental limitation of existing approaches is the lack of interpretability. Many LLM-based fact-checking models function as black-box classifiers, predicting whether a claim is true or false without explicitly showing the reasoning steps. This lack of transparency makes it difficult to verify results, particularly in high-stakes applications such as legal and financial audits~\cite{eisenschlos2020understanding}. Ideally, a fact-checking system should provide a structured reasoning process that can be independently validated.

To address these challenges, we propose a novel approach that reformulates tabular fact verification and question-answering as a structured representation learning task. Rather than assuming LLMs inherently understand tabular structures, we task them to construct explicit reasoning steps as executable \texttt{pandas} queries. Since \texttt{pandas} queries are designed for tabular operations (e.g., filtering, counting, or aggregating), they provide transparent and interpretable reasoning steps on how the answer is obtained.

To train our model, we construct \textit{\tabfact}, an augmented fact-checking dataset based on TabFact. Using the DeepSeek-Chat model~\cite{guo2025deepseek}, we translate TabFact statements into corresponding \texttt{pandas} queries, explicitly encoding the logical reasoning required for verification. We further refine \textit{\tabfact} through automated error correction to ensure syntactical validity and execution robustness. We then fine-tune the DeepSeek-coder-7B-instruct-v1.5~\cite{guo2025deepseek} model on \textit{\tabfact}, effectively distilling structured reasoning from DeepSeek-Chat, which has 671B parameters. Despite the reduced scale, our 7B-parameter model achieves on-par performance with DeepSeek-Chat in fact verification and strong generalization to unseen tabular structures.

To evaluate the generalization ability of our method, we conduct OOD experiments on \textit{\wikifact}, a dataset we derived from WikiTableQuestions~\cite{pasupat2015compositional}. Since this dataset is originally designed for question answering, we convert question-answer pairs in the test set into fact-checking claims to match our verification setup. Without any additional fine-tuning on WikiTableQuestions, our model, trained on \textit{\tabfact}, achieves \textbf{84.72\%} accuracy on \textit{\wikifact}, exhibiting strong robustness to unseen tabular formats and domain shifts, significantly outperforming all other baselines.

Beyond fact verification, we extend our structured approach to tabular question answering, where the goal is to extract precise answers rather than classify claims as true or false. To achieve this, we construct \textit{\wikitable}, a dataset derived from WikiTableQuestions, by converting each question into a corresponding \texttt{pandas} query using DeepSeek-Chat. This dataset consists of 1,200 training examples, ensuring each query correctly retrieves the expected answer from the table. We fine-tune the DeepSeek-coder-7B-instruct-v1.5 model on \textit{\wikitable} and evaluate it on the WikiTableQuestions test set, achieving \textbf{75.1\%} accuracy, comparable to state-of-the-art methods despite the small size of the training data. This demonstrates the broader potential of structured representation learning for tabular reasoning, extending its utility beyond fact verification.

\subsection{Contributions and Paper Organization}

Our contributions are as follows:

\begin{itemize}
    \item Execution-Based Fact Verification: We introduce \textit{\methodname}\space(Reason with Pandas), a method that translates natural language claims into executable \texttt{pandas} queries, ensuring \textbf{interpretable} fact verification. Unlike black-box classifiers, \textit{\methodname} explicitly encodes the reasoning process, allowing users to inspect, validate, and debug fact-checking decisions through executable queries (Section~\ref{sec:methodology}).

    \item \textit{\tabfact}: A Structured Fact-Checking Dataset: We construct \textit{\tabfact}, an augmented version of TabFact where each claim is paired with a \texttt{pandas} query generated using DeepSeek-Chat (Section~\ref{sec:methodology}).

    \item Strong OOD Generalization: We derive \textit{\wikifact} by converting question-answer pairs from the WikiTableQuestions test set into factual claims for fact verification. Without any additional fine-tuning, \textit{\methodname} achieves \textbf{84.72\%} accuracy on \textit{\wikifact}, surpassing state-of-the-art methods in out-of-distribution settings (Section~\ref{sec:experiments}).
    
    \item Extending \textit{\methodname} to Question Answering with \textit{\wikitable}: We introduce \textit{\wikitable}, a dataset where questions from WikiTableQuestions are converted into \texttt{pandas} queries for structured question answering. Fine-tuning on \textit{\wikitable}, \textit{\methodname} achieves \textbf{75.1\%} accuracy, demonstrating its applicability beyond fact verification (Section~\ref{sec:experiments}).

\end{itemize}

The rest of this paper is organized as follows: Section~\ref{sec:related_work} reviews prior work. Section~\ref{sec:methodology} details dataset construction, model architecture, and training. Section~\ref{sec:experiments} presents experimental results, and Section~\ref{sec:conclusion} concludes the paper.

\section{Related Work}
\label{sec:related_work}

\subsection{Fact Verification with Tabular Data}

Fact verification over tabular data has been extensively studied, with datasets like TabFact~\cite{chen2019tabfact} serving as key benchmarks for evaluating models' ability to verify claims about structured data. Early methods relied on sequence-based models such as Table-BERT~\cite{chen2019tabfact}, which linearized tables before applying a pre-trained transformer for classification. However, these methods struggled with complex numerical reasoning and lacked interpretability.

More advanced approaches, such as TAPAS~\cite{herzig2020tapas} and TAPEX~\cite{liu2021tapex}, incorporated table-aware pretraining to improve structured data comprehension. TAPAS extended BERT with table-specific positional embeddings, while TAPEX introduced pretraining over table-based tasks, treating table-based reasoning as a weakly supervised semantic parsing task~\cite{yin2020tabert}. However, both TAPAS and TAPEX function as black-box models, making their decision-making process difficult to interpret, as they do not explicitly provide reasoning steps for their fact-checking predictions. PASTA~\cite{gu2022pasta}, focused on sentence-table cloze pretraining, aiming to teach models table operations such as filtering, aggregation, and comparison. While PASTA improves structured reasoning, it relies on a predefined set of operations, which may limit its applicability to more complex or novel table structures that require reasoning beyond these fixed operations.

Our approach differs by explicitly translating claims into executable \texttt{pandas} queries, ensuring transparent and verifiable fact verification. Unlike previous models, our method explicitly encodes reasoning steps, making the verification process both interpretable and executable. Importantly, rather than relying on dataset-specific patterns, our approach focuses on translating claims into structured \texttt{pandas} queries, enabling it to generalize more effectively to unseen tables and diverse tabular formats.

\subsection{Structured Representation Learning for Tables}

Recent research has explored improving structured reasoning by integrating execution-based frameworks. Program-driven methods, such as ProgVGAT~\cite{yang2020program}, employ graph neural networks (GNNs) to capture logical relationships within tables, while ReasTAP~\cite{zhao2022reastap} applies symbolic reasoning to enhance table comprehension. Additionally, models such as StructGPT~\cite{jiang2023structgpt} and Struct-X~\cite{tan2024struct} encode structured data using graph-based attention mechanisms, but often require complex architectures.

In contrast, our method leverages \texttt{pandas} queries, which naturally define structured table operations (e.g., filtering, aggregation, and row selection) and enable direct execution for fact verification. This aligns with trends in tool-augmented reasoning, where models generate structured outputs (such as SQL, Python scripts, or execution traces) to improve interpretability~\cite{yao2023react, wei2022chain}. Furthermore, we evaluate our approach in out-of-distribution (OOD) settings using WikiTableQuestions~\cite{pasupat2015compositional}, demonstrating strong generalization without additional fine-tuning.

\subsection{Question Answering over Tabular Data}

While fact verification focuses on binary classification (entailed vs. refuted), table-based question answering (QA) presents additional challenges, requiring compositional reasoning over structured data~\cite{chen2020hybridqa}. Methods such as TAPEX~\cite{liu2021tapex} model QA as an SQL generation task, while TabLaP~\cite{wang2024accurate} treats LLMs as planning agents, generating Python-based execution plans to improve numerical reasoning.

Chain-of-Table~\cite{wang2024chain} extends Chain-of-Thought prompting to tabular settings, guiding LLMs through step-by-step execution of table transformations. Similarly, SynTQA~\cite{zhang2024syntqa} leverages text-to-SQL conversion for structured QA, but still struggles with interpretable reasoning steps.

Our method extends fact verification to QA by generating executable \texttt{pandas} queries, demonstrating that structured representation learning enhances both fact-checking and QA performance. Despite training on only 1,200 QA pairs, our approach achieves competitive results compared to state-of-the-art QA models, highlighting the effectiveness of structured execution in table-based reasoning.

\section{Method}
\label{sec:methodology}

\subsection{Problem Formulation}

Given a structured table $\mathcal{T}$ and a natural language statement $s$, the goal of tabular fact verification is to determine whether $s$ is \textit{entailed} or \textit{refuted} based on $\mathcal{T}$. Instead of directly classifying statements, we introduce a structured reasoning approach by translating $s$ into an executable \texttt{pandas} query $q_s$. The execution result of $q_s$ on $\mathcal{T}$ provides a verifiable, interpretable decision process for fact verification. 

Formally, we define a function $f_{\theta}$, parameterized by a language model, that maps a statement to a \texttt{pandas} query:

\begin{equation}
    q_s = f_{\theta}(s, \mathcal{T})
\end{equation}

The execution of $q_s$ on $\mathcal{T}$ produces a verification result to classify $s$ as \textit{entailed} or \textit{refuted}.

\subsection{Dataset Construction}

We construct two datasets to train our model for fact verification and question answering: \textit{\tabfact} for fact-checking and \textit{\wikitable} for tabular QA.

\subsubsection{\textit{\tabfact}: A Fact-Checking Dataset}
\textit{\tabfact} is a structured dataset derived from TabFact~\cite{chen2019tabfact}. Since TabFact consists of tables and annotated claims labeled as entailed or refuted but lacks explicit reasoning steps, we augment it with structured queries to enable execution-based verification. Specifically, we use the DeepSeek-Chat model to generate \texttt{pandas} queries corresponding to each claim, ensuring explicit reasoning for fact verification. Details on \textit{\tabfact} can be found in Appendix \ref{sec:appendix_PanTabFact}.

\begin{figure}[t]
    \centering
    \includegraphics[width=1.0\linewidth]{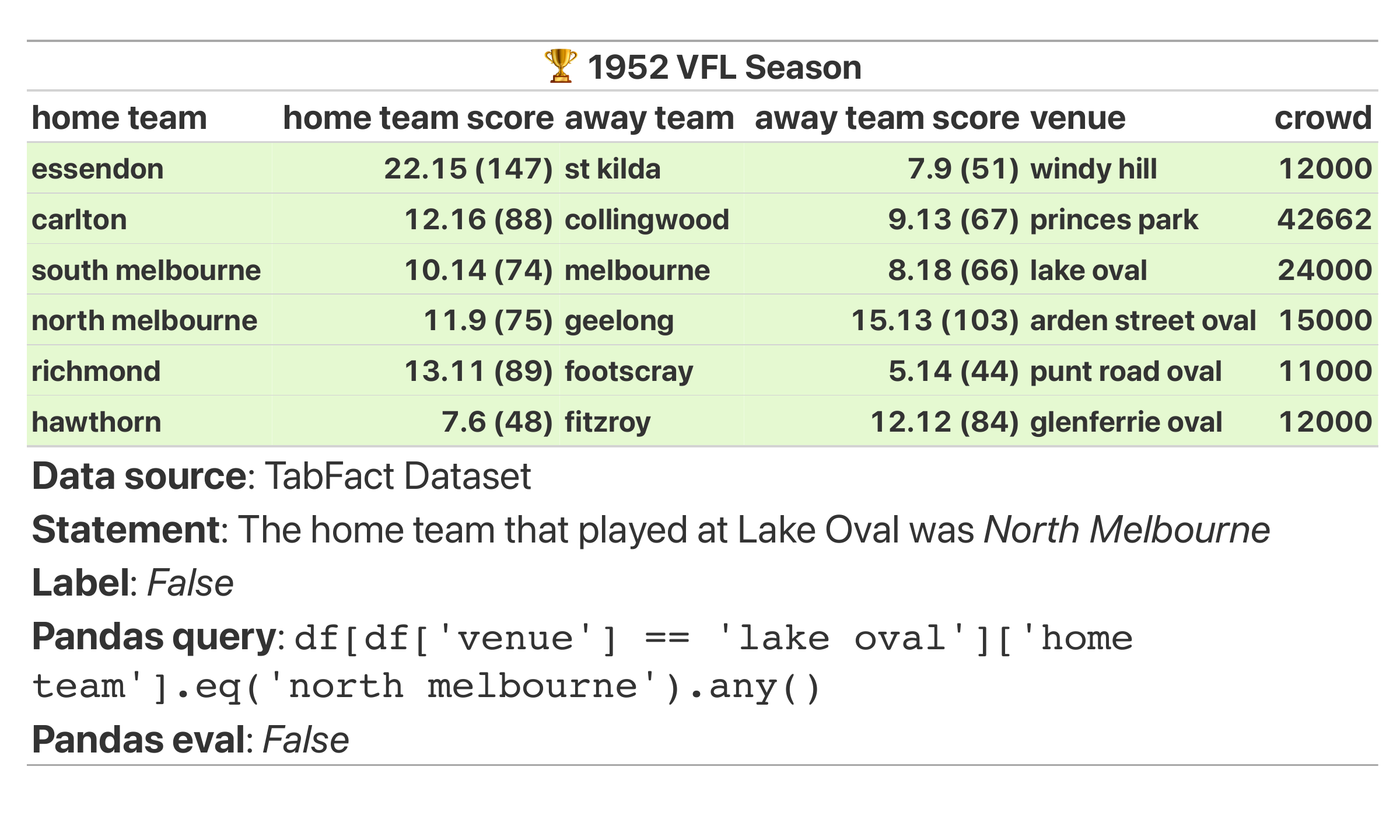}
    \caption{An example from PanTabFact.}
    \label{fig: train_data_example}
\end{figure}

\textbf{Query Generation:} For each statement-table pair $(s, \mathcal{T})$ in TabFact, we prompt DeepSeek-Chat to generate an equivalent \texttt{pandas} query $q_s$. The query should encode the logical operation required to verify $s$ based on $\mathcal{T}$. Figure ~\ref{fig: train_data_example} illustrates an example from \textit{\tabfact}.

\subsubsection{\textit{\wikitable}: A Question-Answering Dataset}
\textit{\wikitable} is a dataset derived from WikiTableQuestions~\cite{pasupat2015compositional} for training \textit{\methodname} in tabular question answering. Each question in WikiTableQuestions is augmented with a \texttt{pandas} query that, when executed, produces the corresponding answer from the dataset. \textit{\wikitable} has 1200 data entries. Details on \textit{\wikitable} can be found in Appendix \ref{sec:appendix_PanWiki}.

\textbf{Query Generation:} For each question-table-answer tuple $(q, \mathcal{T}, a)$ in WikiTableQuestions, we prompt DeepSeek-Chat to generate a \texttt{pandas} query $q_q$ that extracts $a$ when executed on $\mathcal{T}$.

\subsubsection{Error Correction}
Since model-generated queries may contain syntactical or logical errors, the training dataset creation process includes an automated error correction pipeline with three post-processing stages.

\begin{itemize}
    \item \textit{Logic Correction:} Verifies if the execution of the \texttt{pandas} query produces the expected answer. If flawed, we pass the original query and expected outcome back to original model for logical refinement. This stage is only applied in training dataset creation not in the inference phase.
    \item \textit{Syntax Correction:} Iteratively refines queries that fail to execute on $\mathcal{T}$
    due to runtime errors.
    \item \textit{Filtering:} Queries that fail execution or do not match the ground-truth entailment label are removed, ensuring training dataset quality.
\end{itemize}

The Error Correction and Filtering steps are applied in both Fact-Checking and Question-Answering settings.
% h, trim={0cm 1cm 0cm 1cm}, clip

\subsection{Model Framework}
We fine-tune DeepSeek-coder-7B-instruct-v1.5~\cite{guo2025deepseek} to generate \texttt{pandas} queries for both fact verification and question answering. The model is trained autoregressively to generate structured queries that, when executed on a given table, provide verifiable reasoning for fact-checking or directly retrieve answers.

For fact-checking, we fine-tune the model on \textit{\tabfact}, where each claim in TabFact is paired with a corresponding \texttt{pandas} query. The model learns to generate queries that determine whether a claim is entailed or refuted based on the table.

For question answering, we fine-tune the model on \textit{\wikitable}, a dataset derived from WikiTableQuestions where each question is paired with a \texttt{pandas} query that retrieves the correct answer when executed. Unlike fact verification, where queries output Boolean values, here the queries extract the precise answer.

Training optimizes the negative log-likelihood of the correct query:

\begin{equation} \mathcal{L} = - \sum_{t=1}^{T} \log P(q_t | q_{<t}, s, \mathcal{T}; \theta) \end{equation}

where $q_t$ is the $t$-th token in the generated query and $T$ is the query length.

\subsection{Out-of-Distribution (OOD) Generalization}  

To assess the OOD generalization of \textit{\methodname}, we derive \textit{\wikifact}, a fact verification dataset, from WikiTableQuestions~\cite{pasupat2015compositional}. Since WikiTableQuestions is originally designed for question answering, we transform each question-answer pair \((q, \mathcal{T}, a)\) into a factual statement \(s_q\) that asserts \(a\) as the correct answer based on \(\mathcal{T}\). This enables us to evaluate \textit{\methodname} in an unseen fact verification setting without any additional fine-tuning.

\section{Experiments}
\label{sec:experiments}

In this section, we outline our experimental setup, present the key results for fact verification and question answering, and compare \textit{\methodname} with state-of-the-art baselines. We evaluate its performance on both in-distribution (TabFact) and out-of-distribution (\textit{\wikifact}) settings, providing a detailed comparison with existing models in OOD scenarios. Additionally, we assess the effectiveness of \textit{\methodname} in tabular question answering.

\subsection{Experimental Setup}

We fine-tune DeepSeek-coder-7B-instruct-v1.5 on both PanTabFact (fact verification) and PanWiki (question answering). The model is trained in an autoregressive manner to generate \texttt{pandas} queries conditioned on input claims (for fact-checking) or questions (for QA) alongside tabular contexts.

\subsubsection{Training}  
Training is conducted with the TRL library by HuggingFace \citep{vonwerra2022trl}. we use the AdamW optimizer with a learning rate of 2e-4 and cosine learning rate scheduling. We train for 4 epochs with a batch size of 4, applying weight decay to prevent overfitting. Fact verification queries are optimized to generate Boolean outputs, while QA queries are trained to extract precise answers from tables.

\subsubsection{Inference \& Evaluation}  
During inference, the model generates a \texttt{pandas} query for each input, which is then executed to obtain the final verification result (for fact-checking) or extracted answer (for QA). The output is compared against the ground-truth answer:

\paragraph{Fact Verification Accuracy:}  
\begin{equation}
y = \mathbb{I} \left( f_{\theta}(s, \mathcal{T}) = GT \right)
\end{equation}  
where \( y \) is the correctness indicator, \( f_{\theta} \) is the trained model, and \( GT \) is the expected Boolean label.

\paragraph{Question Answering Accuracy:}  
\begin{equation}
y = \mathbb{I} \left( f_{\theta}(q, \mathcal{T}) \approx a \right)
\end{equation}  
where \( q \) is the input question and \( a \) is the ground-truth answer.

Furthermore, we apply Syntax Correction at inference. If a query fails due to syntax errors, we pass the error message back into  DeepSeek-coder-7B-instruct-v1.5, prompting $4$ iterative refinements until a valid, executable query is obtained.

This setup allows \textit{\methodname} to perform structured verification and answer extraction across both in-distribution and out-of-distribution tabular data.

\subsection{In-Distribution Evaluation on PanTabFact}

We first evaluate \textit{\methodname} on the TabFact test set to assess its in-distribution performance. We compare \textit{\methodname}, which generates structured \texttt{pandas} queries for fact verification, against several baselines to evaluate the effectiveness of execution-based reasoning. The baselines include:

% \textbf{\textit{\methodname} (Finetuned-Pandas)}: \textit{\methodname}, fine-tuned on \textit{\tabfact} to generate \texttt{pandas} queries for fact verification.

\textbf{Finetuned-Direct}:  DeepSeek-coder-7B-instruct-v1.5 fine-tuned to classify statements as entailed or refuted directly, without generating \texttt{pandas} queries.

\textbf{ZeroShot-Pandas}: A zero-shot DeepSeek-coder-7B-instruct-v1.5 model that generates \texttt{pandas} queries without fine-tuning.

\textbf{ZeroShot-Direct}: A zero-shot DeepSeek-coder-7B-instruct-v1.5 model that directly classifies claims as entailed or refuted without structured reasoning.

Table~\ref{tab:tabfact_results} presents the accuracy of each method on the TabFact test set.

\begin{table}[h]
    \centering
    \caption{Fact verification accuracy on the TabFact test set. \textit{\methodname} significantly outperforms baselines, demonstrating the effectiveness of structured representation learning and knowledge transfer.}
    \label{tab:tabfact_results}
    \setlength{\tabcolsep}{5pt} % Adjust column spacing
    \renewcommand{\arraystretch}{1.1} % Adjust row spacing
    \begin{tabular}{l c}
        \toprule
        \textbf{Method} & \textbf{Accuracy (\%)} \\
        \midrule
        \textit{\methodname} (Fact-Checking) & \textbf{84.09} \\
        Finetuned-Direct & 67.85 \\
        ZeroShot-Pandas & 51.82 \\
        ZeroShot-Direct & 50.76 \\
        \bottomrule
    \end{tabular}
\end{table}

\textit{\methodname} achieves \textbf{84.09\%} accuracy, significantly outperforming the direct classification baseline (Finetuned-Direct) by \textbf{16.24\%}. Furthermore, it surpasses the ZeroShot-Direct model, which achieves only \textbf{50.76\%} accuracy—close to random guessing—by a margin of \textbf{33.33\%}. These results highlight the effectiveness of \texttt{pandas}-based structured learning, allowing \textit{\methodname} to learn structured reasoning through execution-based fact verification while maintaining strong accuracy.

The stark contrast with ZeroShot baselines highlights the challenge of verifying tabular claims without fine-tuning, as the base model lacks prior exposure to structured data. \textit{\methodname} improves both accuracy and interpretability by translating claims into executable \texttt{pandas} queries, explicitly encoding the reasoning process. Unlike black-box classifiers, \textit{\methodname} provides a transparent verification pipeline where users can inspect the generated queries to validate the logic behind each decision. This structured approach enables an auditable fact-checking process, allowing errors or misclassifications to be traced back to specific reasoning steps, enhancing trust in the verification process.

\subsection{Out-of-Distribution Generalization}

To evaluate the robustness of \textit{\methodname} beyond in-distribution fact verification, we assess its generalization on out-of-distribution (OOD) tabular data using \textit{\wikifact} dataset. This enables us to test whether \textit{\methodname}, trained only on \textit{\tabfact}, can transfer effectively to an unseen dataset without additional fine-tuning.

\paragraph{Performance on \textit{\wikifact} without further Fine-Tuning.} We evaluate \textit{\methodname} on \textit{\wikifact} without fine-tuning. The model, trained solely on \textit{\tabfact}, is tested directly on the transformed fact verification statements from WikiTableQuestions. Table~\ref{tab:ood_results} presents the accuracy results.

\begin{table}[h]
    \centering
    \caption{Fact verification accuracy on \textit{\wikifact} dataset without further fine-tuning.}
    \label{tab:ood_results}
    \setlength{\tabcolsep}{5pt}
    \renewcommand{\arraystretch}{1.1}
    \begin{tabular}{l c}
        \toprule
        \textbf{Method} & \textbf{Accuracy (\%)} \\
        \midrule
        \textit{\methodname} (Fact-Checking) & \textbf{84.72} \\
        Finetuned-Direct & 74.10 \\
        ZeroShot-Pandas & 59.92 \\
        ZeroShot-Direct & 53.20 \\
        \bottomrule
    \end{tabular}
\end{table}

\textit{\methodname} achieves \textbf{84.72\%} accuracy on \textit{\wikifact}, demonstrating strong generalization despite being trained solely on \textit{\tabfact}. It outperforms Finetuned-Direct (\textbf{74.10\%}) by \textbf{10.62\%} while also offering interpretability over the black-box Finetuned-Direct method. Zero-shot models perform significantly worse, with ZeroShot-Direct at \textbf{53.20\%}, reinforcing the importance of knowledge transfer from DeepSeek-Chat by fine-tuning. This improvement stems from \textit{\methodname}'s ability to learn a structured representation that generalizes beyond specific tabular distributions, allowing it to adapt effectively to unseen tables. Since all examples in \textit{\wikifact} are factually correct, one might argue that \textit{\methodname}'s high accuracy on \textit{\wikifact} stems from the model consistently classifying examples as correct. However, in the next section, we demonstrate that this is not the case. \textit{\methodname} achieves \textbf{87\%} accuracy on the balanced dataset we synthesized.

\paragraph{Comparison with Existing Methods on OOD Data.} To further evaluate OOD generalization, we compare \textit{\methodname} with state-of-the-art tabular fact verification models.

\textbf{TAPEX}~\cite{liu2021tapex}: A table-pretrained model using SQL-based execution.

\textbf{TAPAS}~\cite{herzig2020tapas}: A transformer-based model optimized for table-based classification.

\textbf{PASTA}~\cite{gu2022pasta}: A fact-checking model trained on synthesized sentence-table cloze tasks.
 
For this experiment, we randomly sample 300 instances from \textit{\wikifact}. Since this dataset is derived from question-answer pairs, all statements are originally true based on the provided tables. To introduce refuted claims, we use DeepSeek-Chat to slightly modify each correct statement, altering its content based on the table to generate a factually incorrect version. This results in a balanced dataset of 300 true and 300 false statements, allowing us to evaluate how effectively each model distinguishes between entailed and refuted claims in an OOD setting.

Table~\ref{tab:comparison_sota} reports the accuracy for both the original (all true) and altered (all false) claims.

\begin{table}[h]
    \centering
    \caption{Comparison of fact verification accuracy on 300 original and 300 modified \textit{\wikifact} statements.}
    \label{tab:comparison_sota}
    \setlength{\tabcolsep}{5pt}
    \renewcommand{\arraystretch}{1.1}
    \begin{tabular}{l c c c}
        \toprule
        \textbf{Method} & \textbf{All False} & \textbf{All True} &
        \textbf{Overal}\\
        \midrule
        \textit{\methodname} & \textbf{88.33} & \textbf{85.67} &
        \textbf{87.00}\\
        TAPEX & 41.00 & 59.33 & 50.16 \\
        TAPAS & 55.00 & 65.33 & 60.16 \\
        PASTA & 47.67 & 51.67 & 49.67  \\
        \bottomrule
    \end{tabular}
\end{table}

\textit{\methodname} significantly outperforms prior methods, achieving \textbf{88.33\%} accuracy on the altered statements and \textbf{85.67\%} on the original ones. Compared to TAPEX, which achieves only 41.00\% accuracy on the altered set, our model demonstrates a 47.33 percentage point improvement, highlighting its superior performance in OOD setting. Similarly, TAPAS and PASTA struggle with distinguishing between entailed and refuted statements, reinforcing the benefits of structured query-based reasoning.

\paragraph{\textbf{Comparison with Zero-Shot DeepSeek-Chat:}}
These results suggest that structured reasoning through \texttt{pandas} queries provides a more robust fact verification mechanism, improving both accuracy and generalization to unseen tabular distributions. To further validate that our approach effectively captures structured reasoning, we evaluate the zero-shot performance of the much larger DeepSeek-Chat model (671B parameters) on the same fact verification tasks. As detailed in Appendix~\ref{sec:appendix_zeroshot}, \textit{\methodname} achieves results comparable to this significantly larger model and even surpasses it on TabFact, where \textit{\methodname} reaches \textbf{84.09\%} accuracy compared to \textbf{82.62\%} from DeepSeek-Chat. Similarly, on \textit{\wikifact}, \textit{\methodname} achieves \textbf{84.72\%}, closely matching the zero-shot DeepSeek-Chat performance of \textbf{85.39\%}. These results highlight that our fine-tuned 7B model effectively distills structured reasoning from DeepSeek-Chat (671B) while maintaining efficiency and interpretability, enabling local execution without significant performance trade-offs.

\subsection{Application to Tabular Question Answering}  

To evaluate the broader applicability of our structured query generation approach, we apply \textit{\methodname} to tabular question answering using the WikiTableQuestions dataset. Unlike fact verification, where the goal is to determine whether a claim is true or false, question answering requires extracting precise answers from tables.

We fine-tune DeepSeek-coder-7B-instruct-v1.5 on \textit{\wikitable}, a dataset of 1,200 question-answer pairs from WikiTableQuestions, enriched with \texttt{pandas} queries generated using DeepSeek-Chat. Despite the limited training data, our method achieves performance on par with state-of-the-art models. Table~\ref{tab:qa_comparison} provides a comparative analysis.

\begin{table}[h]
    \centering
    \caption{Comparison of tabular question answering accuracy on WikiTableQuestions. Our model uses only 1,200 training examples, significantly fewer than other methods.} \label{tab:qa_comparison}
    \setlength{\tabcolsep}{5pt}
    \renewcommand{\arraystretch}{1.1}
    \begin{tabular}{l c}
        \toprule
        \textbf{Method} & \textbf{Accuracy (\%)} \\
        \midrule
        TabLaP~\cite{wang2024accurate} & \textbf{76.6} \\
        SynTQA (GPT)\cite{zhang2024syntqa} & \textbf{74.4} \\
        Mix SC\cite{liu2023rethinking} & 73.6 \\
        SynTQA (RF)\cite{zhang2024syntqa} & 71.6 \\
        CABINET\cite{patnaik2024cabinet} & 69.1 \\
        Chain-of-Table~\cite{wang2024chain} & 67.31 \\
        Tab-PoT~\cite{xiao2024efficient} & 66.78 \\
        \midrule
        \textit{\methodname} (Finetuned-Pandas for QA) & \textbf{75.1} \\
        \bottomrule
    \end{tabular}
\end{table}

\textit{\methodname} achieves \textbf{75.1\%} accuracy, performing competitively with models like TabLaP and SynTQA (GPT), despite training on only 1,200 examples. In contrast, most existing approaches rely on significantly larger datasets and task-specific optimizations. These results highlight the potential of structured query generation for table-based QA, demonstrating that a \texttt{pandas}-based execution framework provides a lightweight yet effective approach to reasoning over structured data.

\subsection{Ablation Study: Effect of Error Correction}

To assess the impact of our automated correction pipeline, we conduct an ablation study comparing our full model with a variant that omits error correction. We evaluate performance on TabFact, \textit{\wikifact}, and WikiTableQuestions to quantify how syntax and execution refinements contribute to accuracy.

\paragraph{Setup.} 
Our error correction pipeline in inference consists of a single step:

\begin{itemize}
    \item \textbf{Syntax Correction:} Addresses runtime execution failures by analyzing error messages and iteratively refining the query until a valid execution is obtained.
\end{itemize}

\begin{figure}[h]
    \centering
    \includegraphics[width=1.0\linewidth]{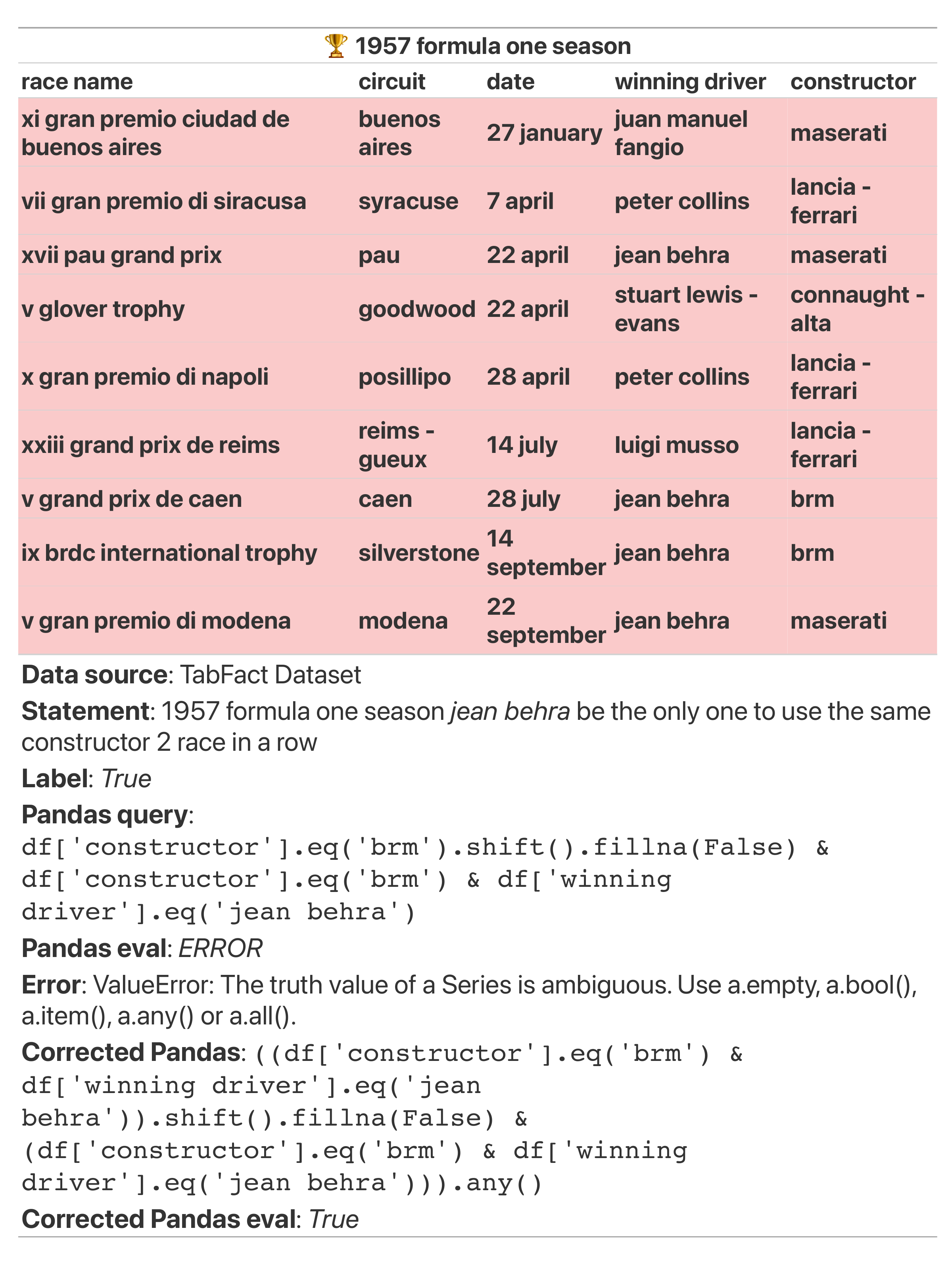}
    \caption{An example of Error Correction.}
    \label{fig: error_correction_example}
\end{figure}

Figure~\ref{fig: error_correction_example} illustrates an example of error correction applied during training dataset creation.

\paragraph{Results.} 
Table~\ref{tab:ablation_error_correction} presents accuracy results with and without corrections modules. Removing error correction results in significant performance degradation across all datasets.

\begin{table}[h]
    \centering
    \caption{Effect of error correction on accuracy in inference time. The absence of error correction leads to a substantial drop in performance.}
    \label{tab:ablation_error_correction}
    \setlength{\tabcolsep}{8pt}
    \renewcommand{\arraystretch}{1.1}
    \begin{tabular}{l c c}
        \toprule
        \textbf{Dataset} & \textbf{No Corr.} & \textbf{With Corr.} \\
        \midrule
        TabFact & 78.02 & \textbf{84.09} \\
        \wikifact & 74.43 & \textbf{84.72} \\
        WQA & 67.59 & \textbf{75.1} \\
        \bottomrule
    \end{tabular}
\end{table}

\paragraph{Analysis.} 
The results emphasize the importance of error correction in structured query generation. Without it, many generated queries fail due to syntax errors. While the underlying logic of the \texttt{pandas} queries is often correct, minor syntax issues—such as missing parentheses or incorrect function calls—can lead to execution failures and misclassifications. Applying error correction significantly enhances reliability by ensuring that structured reasoning remains executable and interpretable.

\section{Conclusion}
\label{sec:conclusion}

We introduced \textit{\methodname}, a structured approach for tabular fact verification that translates claims into executable \texttt{pandas} queries, ensuring interpretable and accurate verification. To support execution-based reasoning, we constructed \textit{\tabfact}, an augmented version of TabFact with structured queries generated via DeepSeek-Chat and refined through automated error correction. Fine-tuning DeepSeek-coder-7B-instruct-v1.5 on \textit{\tabfact}, \textit{\methodname} achieved \textbf{84.09\%} accuracy on TabFact and \textbf{84.72\%} on \textit{\wikifact} without additional fine-tuning, demonstrating strong out-of-distribution (OOD) generalization.

Beyond fact verification, we introduced \textit{\wikitable}, a structured QA dataset with 1200 data entries derived from WikiTableQuestions. Fine-tuning \textit{\methodname} on \textit{\wikitable}, we achieved \textbf{75.1\%} accuracy in table-based QA, showcasing the broader applicability of execution-based reasoning.

Additionally, we compare \textit{\methodname} with the zero-shot DeepSeek-Chat model (671B) on fact verification benchmarks (~\ref{sec:appendix_zeroshot}). Despite the large scale of DeepSeek-Chat, our fine-tuned model achieves comparable or superior accuracy, demonstrating the successful distillation of structured reasoning capabilities into a compact and deployable model.

Unlike black-box classifiers, \textit{\methodname} explicitly encodes reasoning steps through executable \texttt{pandas} queries, ensuring transparent, verifiable, and interpretable fact-checking and question answering. By leveraging structured execution rather than implicit model predictions, \textit{\methodname} enables users to trace and validate the reasoning behind each decision. Its strong performance across diverse tabular distributions demonstrates the effectiveness of execution-based reasoning, setting a new standard for accuracy, generalization, and reliability in tabular fact verification and QA.

\section{Limitations}

One limitation of our work is that we focused solely on fact verification using datasets where each entry consists of a single table. This constraint means our approach has not been evaluated on more complex cases involving multiple tables, cross-table reasoning, or hierarchical data structures. As a result, its effectiveness in scenarios requiring multi-table aggregation or relational inferences remains unexplored. Future work could extend our methodology to handle fact verification across multiple interconnected tables, improving its applicability to real-world datasets with richer relational structures.

\section{Acknowledgements}
This project was supported in part by a grant from an NSF CAREER AWARD 1942230, ONR YIP award N00014-22-1-2271, ARO’s Early Career Program Award 310902-00001, Army Grant No. W911NF2120076, the NSF award CCF2212458, NSF Award No. 2229885 (NSF Institute for Trustworthy AI in Law and Society, TRAILS), a MURI grant 14262683, an award from meta 314593-00001 and an award from Capital One.e
% \clearpage

\bibliography{latex/custom}

\begin{thebibliography}{22}
\providecommand{\natexlab}[1]{#1}

\bibitem[{Chen et~al.(2019)Chen, Wang, Chen, Zhang, Wang, Li, Zhou, and Wang}]{chen2019tabfact}
Wenhu Chen, Hongmin Wang, Jianshu Chen, Yunkai Zhang, Hong Wang, Shiyang Li, Xiyou Zhou, and William~Yang Wang. 2019.
\newblock Tabfact: A large-scale dataset for table-based fact verification.
\newblock \emph{arXiv preprint arXiv:1909.02164}.

\bibitem[{Chen et~al.(2020)Chen, Zha, Chen, Xiong, Wang, and Wang}]{chen2020hybridqa}
Wenhu Chen, Hanwen Zha, Zhiyu Chen, Wenhan Xiong, Hong Wang, and William Wang. 2020.
\newblock Hybridqa: A dataset of multi-hop question answering over tabular and textual data.
\newblock \emph{arXiv preprint arXiv:2004.07347}.

\bibitem[{Eisenschlos et~al.(2020)Eisenschlos, Krichene, and M{\"u}ller}]{eisenschlos2020understanding}
Julian~Martin Eisenschlos, Syrine Krichene, and Thomas M{\"u}ller. 2020.
\newblock Understanding tables with intermediate pre-training.
\newblock \emph{arXiv preprint arXiv:2010.00571}.

\bibitem[{Gu et~al.(2022)Gu, Fan, Tang, Nakov, Zhao, and Du}]{gu2022pasta}
Zihui Gu, Ju~Fan, Nan Tang, Preslav Nakov, Xiaoman Zhao, and Xiaoyong Du. 2022.
\newblock Pasta: table-operations aware fact verification via sentence-table cloze pre-training.
\newblock \emph{arXiv preprint arXiv:2211.02816}.

\bibitem[{Guo et~al.(2025)Guo, Yang, Zhang, Song, Zhang, Xu, Zhu, Ma, Wang, Bi et~al.}]{guo2025deepseek}
Daya Guo, Dejian Yang, Haowei Zhang, Junxiao Song, Ruoyu Zhang, Runxin Xu, Qihao Zhu, Shirong Ma, Peiyi Wang, Xiao Bi, et~al. 2025.
\newblock Deepseek-r1: Incentivizing reasoning capability in llms via reinforcement learning.
\newblock \emph{arXiv preprint arXiv:2501.12948}.

\bibitem[{Herzig et~al.(2020)Herzig, Nowak, M{\"u}ller, Piccinno, and Eisenschlos}]{herzig2020tapas}
Jonathan Herzig, Pawe{\l}~Krzysztof Nowak, Thomas M{\"u}ller, Francesco Piccinno, and Julian~Martin Eisenschlos. 2020.
\newblock Tapas: Weakly supervised table parsing via pre-training.
\newblock \emph{arXiv preprint arXiv:2004.02349}.

\bibitem[{Jiang et~al.(2023)Jiang, Zhou, Dong, Ye, Zhao, and Wen}]{jiang2023structgpt}
Jinhao Jiang, Kun Zhou, Zican Dong, Keming Ye, Wayne~Xin Zhao, and Ji-Rong Wen. 2023.
\newblock Structgpt: A general framework for large language model to reason over structured data.
\newblock \emph{arXiv preprint arXiv:2305.09645}.

\bibitem[{Liu et~al.(2021)Liu, Chen, Guo, Ziyadi, Lin, Chen, and Lou}]{liu2021tapex}
Qian Liu, Bei Chen, Jiaqi Guo, Morteza Ziyadi, Zeqi Lin, Weizhu Chen, and Jian-Guang Lou. 2021.
\newblock Tapex: Table pre-training via learning a neural sql executor.
\newblock \emph{arXiv preprint arXiv:2107.07653}.

\bibitem[{Liu et~al.(2023)Liu, Wang, and Chen}]{liu2023rethinking}
Tianyang Liu, Fei Wang, and Muhao Chen. 2023.
\newblock Rethinking tabular data understanding with large language models.
\newblock \emph{arXiv preprint arXiv:2312.16702}.

\bibitem[{Pasupat and Liang(2015)}]{pasupat2015compositional}
Panupong Pasupat and Percy Liang. 2015.
\newblock Compositional semantic parsing on semi-structured tables.
\newblock \emph{arXiv preprint arXiv:1508.00305}.

\bibitem[{Patnaik et~al.(2024)Patnaik, Changwal, Aggarwal, Bhatia, Kumar, and Krishnamurthy}]{patnaik2024cabinet}
Sohan Patnaik, Heril Changwal, Milan Aggarwal, Sumit Bhatia, Yaman Kumar, and Balaji Krishnamurthy. 2024.
\newblock Cabinet: Content relevance based noise reduction for table question answering.
\newblock \emph{arXiv preprint arXiv:2402.01155}.

\bibitem[{Tan et~al.(2024)Tan, Wang, Qiu, Cheng, Xu, Chu, and Qi}]{tan2024struct}
Xiaoyu Tan, Haoyu Wang, Xihe Qiu, Yuan Cheng, Yinghui Xu, Wei Chu, and Yuan Qi. 2024.
\newblock Struct-x: Enhancing large language models reasoning with structured data.
\newblock \emph{arXiv preprint arXiv:2407.12522}.

\bibitem[{von Werra et~al.(2020)von Werra, Belkada, Tunstall, Beeching, Thrush, Lambert, Huang, Rasul, and Gallouédec}]{vonwerra2022trl}
Leandro von Werra, Younes Belkada, Lewis Tunstall, Edward Beeching, Tristan Thrush, Nathan Lambert, Shengyi Huang, Kashif Rasul, and Quentin Gallouédec. 2020.
\newblock Trl: Transformer reinforcement learning.
\newblock \url{https://github.com/huggingface/trl}.

\bibitem[{Wang et~al.(2024{\natexlab{a}})Wang, Qi, and Gan}]{wang2024accurate}
Yuxiang Wang, Jianzhong Qi, and Junhao Gan. 2024{\natexlab{a}}.
\newblock Accurate and regret-aware numerical problem solver for tabular question answering.
\newblock \emph{arXiv preprint arXiv:2410.12846}.

\bibitem[{Wang et~al.(2024{\natexlab{b}})Wang, Zhang, Li, Eisenschlos, Perot, Wang, Miculicich, Fujii, Shang, Lee et~al.}]{wang2024chain}
Zilong Wang, Hao Zhang, Chun-Liang Li, Julian~Martin Eisenschlos, Vincent Perot, Zifeng Wang, Lesly Miculicich, Yasuhisa Fujii, Jingbo Shang, Chen-Yu Lee, et~al. 2024{\natexlab{b}}.
\newblock Chain-of-table: Evolving tables in the reasoning chain for table understanding.
\newblock \emph{arXiv preprint arXiv:2401.04398}.

\bibitem[{Wei et~al.(2022)Wei, Wang, Schuurmans, Bosma, Xia, Chi, Le, Zhou et~al.}]{wei2022chain}
Jason Wei, Xuezhi Wang, Dale Schuurmans, Maarten Bosma, Fei Xia, Ed~Chi, Quoc~V Le, Denny Zhou, et~al. 2022.
\newblock Chain-of-thought prompting elicits reasoning in large language models.
\newblock \emph{Advances in neural information processing systems}, 35:24824--24837.

\bibitem[{Xiao et~al.(2024)Xiao, Kantarci, Kang, Niyato, and Guizani}]{xiao2024efficient}
Bin Xiao, Burak Kantarci, Jiawen Kang, Dusit Niyato, and Mohsen Guizani. 2024.
\newblock Efficient prompting for llm-based generative internet of things.
\newblock \emph{arXiv preprint arXiv:2406.10382}.

\bibitem[{Yang et~al.(2020)Yang, Nie, Feng, Liu, Chen, and Zhu}]{yang2020program}
Xiaoyu Yang, Feng Nie, Yufei Feng, Quan Liu, Zhigang Chen, and Xiaodan Zhu. 2020.
\newblock Program enhanced fact verification with verbalization and graph attention network.
\newblock \emph{arXiv preprint arXiv:2010.03084}.

\bibitem[{Yao et~al.(2023)Yao, Zhao, Yu, Du, Shafran, Narasimhan, and Cao}]{yao2023react}
Shunyu Yao, Jeffrey Zhao, Dian Yu, Nan Du, Izhak Shafran, Karthik Narasimhan, and Yuan Cao. 2023.
\newblock React: synergizing reasoning and acting in language models (2022).
\newblock \emph{arXiv preprint arXiv:2210.03629}.

\bibitem[{Yin et~al.(2020)Yin, Neubig, Yih, and Riedel}]{yin2020tabert}
Pengcheng Yin, Graham Neubig, Wen-tau Yih, and Sebastian Riedel. 2020.
\newblock Tabert: Pretraining for joint understanding of textual and tabular data.
\newblock \emph{arXiv preprint arXiv:2005.08314}.

\bibitem[{Zhang et~al.(2024)Zhang, Luu, and Zhao}]{zhang2024syntqa}
Siyue Zhang, Anh~Tuan Luu, and Chen Zhao. 2024.
\newblock Syntqa: Synergistic table-based question answering via mixture of text-to-sql and e2e tqa.
\newblock \emph{arXiv preprint arXiv:2409.16682}.

\bibitem[{Zhao et~al.(2022)Zhao, Nan, Qi, Zhang, and Radev}]{zhao2022reastap}
Yilun Zhao, Linyong Nan, Zhenting Qi, Rui Zhang, and Dragomir Radev. 2022.
\newblock Reastap: Injecting table reasoning skills during pre-training via synthetic reasoning examples.
\newblock \emph{arXiv preprint arXiv:2210.12374}.

\end{thebibliography}

\clearpage
\appendix

\section{Appendix}
\label{sec:appendix}

\subsection{Training Dataset Creation}
\label{sec:appendix_training_data}

\subsubsection{PanTabFact}
\label{sec:appendix_PanTabFact}
To construct the training dataset, we generate pandas queries for statements in TabFact using DeepSeek-Chat. The dataset undergoes multiple correction phases to improve syntax, and logical accuracy. Table~\ref{tab:dataset_creation} summarizes the statistics at each stage.

\begin{table}[!h]
    \centering
    \caption{\methodname\ statistics across different correction phases. Accuracy represents the proportion of correctly classified executable queries.}
    \label{tab:dataset_creation}
    \setlength{\tabcolsep}{8pt}
    \renewcommand{\arraystretch}{1.1}
    \begin{tabular}{l c c}
        \toprule
        \textbf{Phase} & \textbf{Correct} & \textbf{Accuracy (\%)} \\
        \midrule
        Initial Generation & 73,172 & 79.29 \\
        % Format Correction & 73,917 & 80.09 \\
        Logic Correction & 84,023 & 91.05 \\
        Syntax Correction & 88,299 & 95.68 \\
        \bottomrule
    \end{tabular}
\end{table}

The initial generation phase produces many syntax errors. Logic correction refines logical inconsistencies before execution. In addition, syntax correction resolves execution failures, resulting in 88,299 valid queries (95.68\% of the original TabFact dataset) in the final dataset. The prompts used for every stage of training dataset creation can be found in Table \ref{tab:training_prompts}.

\subsubsection{PanWikiQA}
\label{sec:appendix_PanWiki}
To construct the question-answering training dataset, we generate pandas queries for WikiTableQuestions using DeepSeek-Chat. The dataset consists of 1,200 training examples created with the instruction prompt in Table~\ref{tab:qa_training_prompt}. The correctness of generated pandas queries is determined by whether their execution produces the exact answer given in the WikiTableQuestions dataset.

Unlike the fact-checking dataset, we did not apply any correction modules in this setting, as our goal was only to showcase that the method is also effective for question answering.

\subsection{Zero-Shot Performance of DeepSeek-Chat}
\label{sec:appendix_zeroshot}

To assess the baseline performance of a larger instruction-tuned model in a \texttt{pandas}-based setting, we evaluated DeepSeek-Chat (zero-shot) on both TabFact and \textit{\wikifact} in fact verification setting. The model was prompted to generate \texttt{pandas} queries corresponding to given claims, using the format outlined in Table~\ref{tab:qa_training_prompt} for both datasets. Since this is an inference-time evaluation, the Correct Logic module was not applied. The results of this zero-shot experiment are presented in Table~\ref{tab:zeroshot_results}.

\begin{table}[h]
    \centering
    \caption{Zero-shot accuracy of DeepSeek-Chat on TabFact and \textit{\wikifact} testsets in fact verification setting, before and after error correction.}
    \label{tab:zeroshot_results}
    \setlength{\tabcolsep}{8pt}
    \renewcommand{\arraystretch}{1.1}
    \begin{tabular}{l c c}
        \toprule
        \textbf{Dataset} & \textbf{No Corr.} & \textbf{With Corr.} \\
        \midrule
        TabFact & 73.38 & \textbf{82.62} \\
        WikiFact & 78.23 & \textbf{85.39} \\
        \bottomrule
    \end{tabular}
\end{table}

\paragraph{Analysis.} The results in Table~\ref{tab:zeroshot_results} show that DeepSeek-Chat, a 671B parameter instruction-tuned model, demonstrates strong zero-shot fact verification capabilities when prompted to generate \texttt{pandas} queries. Notably, after applying error correction, its accuracy improves significantly, highlighting the importance of structured execution refinement. Since our training data is derived from this large model, the fact that \textit{\methodname} achieves similar performance—and even surpasses it in the case of TabFact (84.09\%)—indicates that our fine-tuning approach effectively transfers structured reasoning knowledge into a much smaller model. Furthermore, as shown in Table~\ref{tab:ood_results}, DeepSeek-7B achieves only 59.92\% accuracy when tested zero-shot on \textit{\wikifact}, whereas \textit{\methodname} reaches 84.72\% without any fine-tuning on this dataset. This demonstrates that knowledge transfer from DeepSeek-Chat significantly enhances the structured reasoning ability of our smaller model, enabling it to generalize effectively to unseen tabular distributions.

\begin{table*}[h]
    \centering
    \caption{\textit{Prompts used for generation and refinement of \tabfact.}}
    \label{tab:training_prompts}
    \renewcommand{\arraystretch}{1.2}
    \begin{tabular}{p{3.5cm} p{12cm}}
        \midrule
        Generation & You are a Python expert specializing in pandas. Your task is to translate the given natural language statement into a single-line pandas expression. 
This expression must be valid and executable to verify the truth of the statement using the provided table.
Consider the following:

1. The table is represented as a pandas DataFrame named df.

2. Do not include explanations, comments, or multiline outputs.

3. Ensure the output is concise, correct, and when run outputs either True or False, and strictly in the following Json Format with a single key "PANDA":
{"PANDA": "<your Pandas code>"} \\
        \midrule
        Correct Logic & You are an expert in Python with a specialization in pandas. Your task is to verify and correct a given pandas code that translates a natural language statement into a pandas expression. The corrected pandas code must accurately evaluate the truth of the statement when applied to the given table.
Requirements:

1. The table is represented as a pandas DataFrame named df.

2. The pandas code must evaluate to a boolean value (True or False) using the snippet: str(bool(eval(pandas\_code))).

3. The corrected pandas code should match the truth value indicated by the provided "Label".

4. Ensure the output is concise, correct, and when run outputs either True or False, and strictly in the following Json Format with a single key "CORRECT PANDA":
{"CORRECT PANDA": "<your Pandas code>"}\\
        \midrule
        Correct Syntax & You are a Python expert specializing in pandas. Your task is to correct a pandas code that translates a given natural language statement into a pandas expression. The code, along with the specific error it contains, is provided. Your corrected pandas\_code must be valid and executable by running the code snippet str(bool(eval(pandas\_code))) ensuring it accurately evaluates the truth of the statement using the provided table with no errors.
        
Make sure the pandas\_code is of type boolean.
Consider the following:

1. The table is represented as a pandas DataFrame named df.

2. Do not include explanations, comments, or multiline outputs.

3. Ensure the output is concise, correct, and when run outputs either True or False, and strictly in the following Json Format with a single key "CORRECT PANDA":
{"CORRECT PANDA": "<your Pandas code>"}\\
        \bottomrule
    \end{tabular}
\end{table*}

\begin{table*}[t!]
    \centering
    \caption{\textit{Prompt used for generating \wikitable.}}
    \label{tab:qa_training_prompt}
    \renewcommand{\arraystretch}{1.2}
    \begin{tabular}{p{3.5cm} p{12cm}}
        \toprule
        \textbf{Task} & \textbf{Prompt} \\
        \midrule
        Generation & You are a Python expert specializing in pandas. You are given a table, a question, and an answer. Your task is to translate the given natural language question into a single-line pandas expression. 
This expression, which acts like a query, must be valid and executable so that running the pandas expression will output the answer to the question.
Consider the following:

1. The table is represented as a pandas DataFrame named df.

2. Do not include explanations, comments, or multiline outputs.

3. Ensure the output is concise, correct, and when run, it outputs the correct given answer, and strictly follows the Json format:
\{"PANDA": "<your Pandas code>"\} \\
        \bottomrule
    \end{tabular}
\end{table*}

\end{document}